\documentclass{article}

\usepackage{graphicx}
\usepackage{subcaption}
\usepackage{amsmath}
\usepackage[table]{xcolor}
\usepackage{multirow}
\usepackage{booktabs}
\usepackage{array}

 \usepackage[ELLISworkshop, final]{neurips_2025}

\usepackage[utf8]{inputenc} 
\usepackage[T1]{fontenc}    
\usepackage{hyperref}       
\usepackage{url}            
\usepackage{booktabs}       
\usepackage{amsfonts}       
\usepackage{nicefrac}       
\usepackage{microtype}      
\usepackage{xcolor}         

\title{
Bridging Chemists and AI: An Expert-Augmented Framework for Interpretable Route Evaluation
}
\workshoptitle{Generative Models, LLMs, and the Future of Molecular AI (ML4Molecules 2025)}

\author{
    Yujia~Guo$^{1\ast}$, \And
    Mikhail~Kabeshov$^{2\ast}$,\And
    Tat~Hong~Duong~Le$^{1}$,\And
    Samuel~Genheden$^{2}$,\And
    Marco~V.~Mijangos$^{2}$,\And
    Varvara~Voinarvoska$^{2,3}$,\And
    Giulia~Bergonzini$^{2}$,\And
    Ola~Engkvist$^{2,3}$,\And
    Samuel~Kaski$^{1,4}$ \\[2mm]
    \small$^{1}$Department of Computer Science, Aalto University, Espoo, Finland.\and
    \small$^{2}$Discovery Sciences R\&D, AstraZeneca, Gothenburg, Sweden.\and
    \small$^{3}$Department of Computer Science and Engineering, \and \small{Chalmers University of Technology and University of Gothenburg,  Gothenburg, Sweden}.\and
    \small$^{4}$Department of Computer Science, University of Manchester, Manchester, UK.\and
    \small$^\ast$Corresponding authors: \texttt{yujia.guo@aalto.fi}, \texttt{mikhail.kabeshov@astrazeneca.com}
}

\begin{document}

\maketitle

\begin{abstract}

Selecting efficient multi-step synthetic routes is a central challenge in organic synthesis, particularly in medicinal and process chemistry, where route choice directly impacts feasibility, cost, and development efficiency.
The data-driven assessement systems often oversimplify the multi-objective nature of synthesis design and rely on proxy datasets (e.g., patent routes) rather than universal grounded criteria. To address this, we introduce an expert-augmented, data-driven scoring framework that integrates machine learning with chemists’ domain knowledge for both numerical and explainable route assessment. A DeepSets-based model is trained using tree edit distance between reference and machine-generated routes, then fine-tuned with expert evaluations to produce both quantitative scores and interpretable qualitative categories (Good, Plausible, Bad). The resulting system achieves an Spearman correlation coefficient of 0.78 ± 0.05 and Pearson correlation of 0.77 ± 0.06 regarding its category assessment prediction, and 60.2\% top-1 ranking accuracy regarding its score prediction, substantially outperforming the previous baseline (17.5\%).

\end{abstract}

\vspace{-1mm}
\section{Introduction}
\vspace{-1mm}

The selection of an appropriate multi-step synthetic route is crucial in organic synthesis, particularly in medicinal and process chemistry, where it plays a central role in drug discovery and development \cite{blacker2011pharmaceutical}. 
While Computer-Aided Synthesis Planning (CASP) tools have achieved success in generating diverse route options, the subsequent evaluation of these routes remains a critical bottleneck.
Although substantial efforts have been made to improve single-step retrosynthetic transformations, the feasibility prediction of forward synthesis remains challenging, partly due to the lack of high-quality negative reaction data~\cite{maloney2023negative}. Furthermore, the focus on high "success rates" in CASP can be misleading if the underlying single-step model may be overly permissive to allow unrealistic retrosynthetic steps~\cite{maziarz2025re}. 
The multi-objective nature of synthesis planning also needs expert to balance off, where factors such as feasibility, cost, route length, availability of starting materials, sustainability, safety, and scalability often conflict. 
Consequently, the field must exert domain knowledge to filter these options, relying heavily on the expertise of synthetic chemists to answer the ultimate chemical question: Is this route experimentally feasible and optimal?

Currently, this reliance on human expertise creates a dependency on manual evaluation, which is slow, expensive, and difficult to scale~\cite{stevens2011progress}. To automate this, recent machine learning methods have treated patent-extracted routes as proxies for "expert-validated" ground truth, training models to distinguish these from machine-generated routes~\cite{mo2021evaluating,li2024retro}. However, patents often reflect time-constrained choices rather than optimal chemistry. The existing datasets is sparse~\cite{maziarz2025re} because usually only one or a few possible syntheses have been reported for most molecules. Since retrosynthesis is multi-objective and lacks a universal objective truth, optimizing solely on patent data risks overfitting to the appearance of a patent route rather than learning the underlying reasoning and evaluation heuristics used by chemists.
\vspace{-1mm}
\begin{figure}
    \centering
    \includegraphics[width=0.4\linewidth]{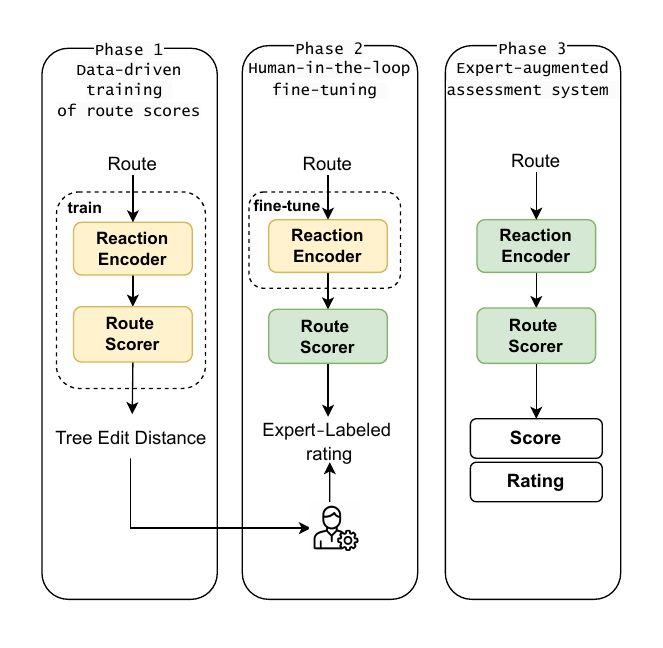}
    \caption{\textbf{Pipeline for developing an expert-augmented deep learning system.} 
    }
    \label{fig:pipline}
    \vspace{-1mm}
\end{figure}

In this work, we propose an expert-augmented framework (Figure~\ref{fig:pipline}) that combines large-scale data-driven modeling with small-scale, expert-aligned assessment, providing interpretable metrics actionable in synthesis practice. We aim to learn a generalizable model capable of both scoring and rating synthetic routes, predicting feasibility based on intrinsic chemical merit while capturing how experts resolve multi-objective trade-offs.
To achieve this, we first employ a similarity comparison metric to construct a DeepSets-based~\cite{zaheer2017deep} scoring model designed with multiple route-level objectives and reaction features, trained on a large corpus of patent routes. 
Then, to bridge the gap between numerical similarity and human judgment, we then design a qualitative reaction-to-route evaluation criterion that translates complex chemical reasoning into a learnable framework. 
We subsequently employ a fine-tuning strategy to adapt the model using expert assessments derived from this specific criterion. This allows the model to learn a distinct route feasibility rating, efficiently leveraging limited but high-value human assessment data. 
Our expert-augmented framework provides a more reliable, transparent, and dual-output (multi-objective quality score + feasibility rating) evaluation of synthetic routes, better aligned with real-world decision-making in retrosynthetic planning.

In quantitative ranking, The model reached 60.2\% top-1 accuracy and an RMSE of 2.72 using SDF embeddings on the test dataset, outperforming previous baselines~\cite{mo2021evaluating} (17\% vs. 60\% top-1 accuracy). The model fine-tuned on this expert-labeled framework, is able to predict not just a numerical similarity score, but an interpretable qualitative evaluation of feasibility. The model achieved a Spearman correlation of $0.782 \pm 0.050$, a Pearson correlation of $ 0.769 \pm 0.064$ with expert ratings in 5-point prediction, and $67 \pm 6.4\%$ classification accuracy in the final three-tier rating, demonstrating its ability to successfully capture the nuance of expert chemical judgment.

\vspace{-1mm}

\section{Expert-augmented route assessment system}

\vspace{-1mm}

\subsection{Pre-training of DeepSets-based scoring model}

We constructed the evaluation model using the \textit{DeepSets}~\cite{zaheer2017deep}neural network architecture, incorporating an embedding of both the route and the route property values. It is specifically designed to process sets with permutation-invariance. This allowed the model to robustly handle varying numbers of reactions and unordered inputs, ensuring flexibility and scalability in route evaluation.
This model predicts the TED score of any synthetic route without knowing the reference route. The information of all reactions in one route is processed as route encoding, and then combined with other route features such as complexity and cost to predict the final TED score. Through the manuscript, we refer to the reaction-level processing network as the \textbf{encoder} network, and the route score prediction network as the \textbf{scorer} network.

Formally, let \(X = \{\mathbf{x}_1, \mathbf{x}_2, \dots, \mathbf{x}_n\}\) represent the set of reaction features of $n$ reactions in one route.
Each reaction \(i\) is represented by a feature vector \(\mathbf{x}_i = \{x^1_i, x^2_i, x^3_i, \mathbf{x}^4_i\}\), where the components include the \textbf{reaction features} detailed in Section~\ref{section:feature}: the reaction class, feasibility score, target compound representation vector, and an optional representation vector (if route embedding is employed).
The encoder \(\phi\) we built satisfies
\begin{equation}
    \phi(\mathbf{x}_1, \mathbf{x}_2, \dots, \mathbf{x}_n) = \phi(\mathbf{x}_{\pi(1)}, \mathbf{x}_{\pi(2)}, \dots, \mathbf{x}_{\pi(n)}),
\end{equation}
where \(\pi\) is a permutation function that reorders these elements. Regardless of how the order of reactions is shuffled by $\pi$, the function \(\phi\) produces the same encoding. This ensures that the output remains the same, regardless of how the input reactions are ordered.

Then, the aggregated reaction encoding from the encoder was concatenated with the route properties, including cost, price, and complexity, along with the molecule fingerprints, before being passed to the scorer network to compute the TED score. 
The fingerprint used as input in both networks is designed to generate target-specific reaction encodings and target-specific route scoring. This setup also allowed the two models to function independently, accommodating various practical demands or constraints. The complete model can be expressed as:
\begin{align}
\text{SCORE} &= \sigma \left( \phi(\mathbf{x}_1, \mathbf{x}_2, \dots, \mathbf{x}_n), c, e , v \right) \\
                &= \sigma \left( \sum_{i=1}^{n} f(\mathbf{x}_i), c, e, v \right)
\end{align}
where \(f\) is a neural network applied to each reaction feature list, and the transformed inputs are summed, producing a representation that is invariant to the reaction order. The scorer net \(\sigma\) processes the aggregated reaction encoding as well as the \textbf{route features} cost $c$, volume $v$ and complexity $e$.

In summary, as shown in Figure~\ref{fig:scoring_model}, the DeepSets-based network architecture consists of two primary modules: the \textbf{reaction-level encoder $\phi$} and \textbf{route-level scorer $\sigma$}, which take in route-level properties and reaction-level features as their inputs, respectively. The encoder network processes each reaction individually, producing a route-level encoding that, along with additional route properties, is passed to the scorer to compute the final score.

\subsection{Fine-tuning with expert assessment}

To validate the score predictions of our DeepSets-based model and align the feasibility prediction to expert assessment, we establish reaction-to-route assessment heuristics, and expert chemists performed a systematic analysis of the route quality as well as suggested scores from the prior model according totheses heuristics. 

First, we established a clear three-tier reaction-to-route classification system with validation criteria for route assessment. 
These metrics ought to produce a completely trackable scoring system, assigning a feasibility value between 1 to 5 points to each individual step of the route. 
Heuristic scoring metrics used by experts to assess route quality are summarized in Table~\ref{tab:reaction_quality_heuristics}. 
All individual scoring assessments are rigorously based in sound literature precedent. 
For the complete synthetic scheme score, we explored two different approaches:  1. \textbf{averaging the step feasibility points}, and 2. using the value of the step with the \textbf{lowest feasibility} as a representative score of the complete route. Finally, we established a three-tier classification system: "good", "plausible", "bad" by merging 3, 4 points and 1, 2 points for interpretable route assessment.

\vspace{-1mm}
\begin{table}[h!]
\centering
\scriptsize
\caption{\textbf{Heuristics for reaction quality assessment.}}
\label{tab:reaction_quality_heuristics}
\renewcommand{\arraystretch}{1.25}
\setlength{\tabcolsep}{5pt}
\begin{tabular}{>{\centering\arraybackslash}p{1cm} >{\centering\arraybackslash}p{1cm} p{8cm}}
\toprule
\textbf{Rating} & \textbf{Points} & \textbf{Criteria} \\
\midrule
\cellcolor{green!20} & \cellcolor{green!20} & \cellcolor{green!20} a) Reaction is highly likely to succeed.\\
\cellcolor{green!20} & \cellcolor{green!20} & \cellcolor{green!20} b) Functional groups have negligible impact on reactivity or selectivity.\\
\multirow{-3}{*}{\cellcolor{green!20}\textbf{good}}
& \multirow{-3}{*}{\cellcolor{green!20}\textbf{5}}
& \cellcolor{green!20} c) Good to moderate regio-/chemoselectivity expected.\\[2pt]

\cellcolor{green!8} & \cellcolor{green!8} & \cellcolor{green!8} a) Disconnection is correct but requires minor substrate or reagent adjustment for optimal reactivity.\\
\multirow{-5}{*}[-7.5ex]{\cellcolor{green!8}\textbf{plausible}} 
& \multirow{-2}{*}{\cellcolor{green!8}\textbf{4}}
& \cellcolor{green!8} b) Includes unnecessary yet retrosynthetically valid steps.\\[2pt]

\cellcolor{green!8} & \cellcolor{green!8} & \cellcolor{green!8} a) Retrosynthetically correct with a suitable protecting-group scheme.\\
\cellcolor{green!8} & \cellcolor{green!8} & \cellcolor{green!8} b) Protection/deprotection or redox maneuver enables transformation.\\
\cellcolor{green!8} 
& \multirow{-3}{*}{\cellcolor{green!8}\textbf{3}}
& \cellcolor{green!8} c) Applies to inadequate protecting group choices.\\[2pt]

\cellcolor{red!15} & \cellcolor{red!15} & \cellcolor{red!15} a) Major regio- or chemoselectivity issues.\\
\multirow{-4}{*}[-6ex]{\cellcolor{red!15}\textbf{bad}} 
& \multirow{-2}{*}{\cellcolor{red!15}\textbf{2}}
& \cellcolor{red!15} b) Implementation of protection approach would alter the route conceptually.\\[2pt]

\cellcolor{red!15} & \cellcolor{red!15} & \cellcolor{red!15} a) Reaction is highly unlikely to succeed.\\
\cellcolor{red!15} 
& \multirow{-2}{*}{\cellcolor{red!15}\textbf{1}}
& \cellcolor{red!15} b) Unresolved to commercially available starting materials.\\
\bottomrule
\end{tabular}
\end{table}

\begin{figure*}[htbp]
\centering

\begin{minipage}{0.48\linewidth}
    \centering
    \includegraphics[width=\linewidth]{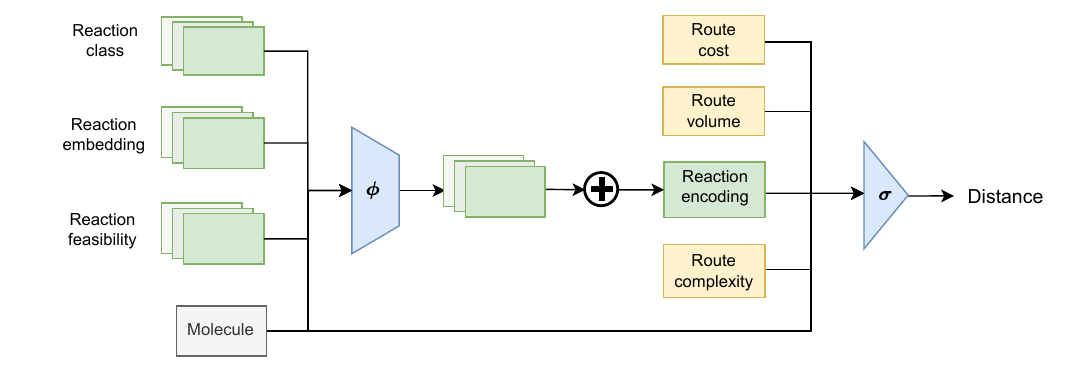}
    \caption{\textbf{Pre-trained scoring model architecture.}  Reaction encoder $\phi$ and scorer net $\sigma$ are trained on offline data.
    }
    \label{fig:scoring_model}
\end{minipage}
\hfill
\begin{minipage}{0.48\linewidth}
    \centering
    \includegraphics[width=\linewidth]{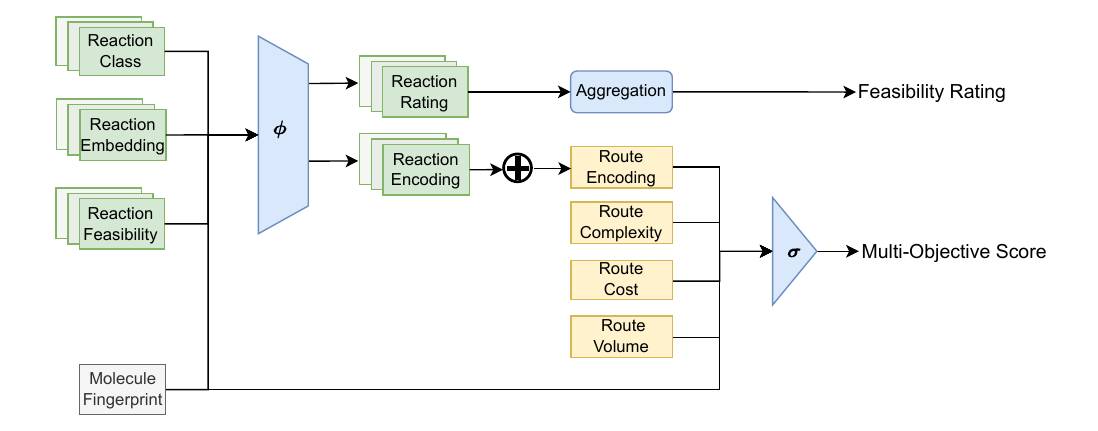}
    \caption{\textbf{Fine-tuned model architecture.} 
    The reaction encoder $\phi$ is fine-tuned using expert-labeled data. Reaction points are aggregated into a route-level feasibility rating.}
    \label{fig:fine_tuned_model}
\end{minipage}
\end{figure*}

Using these criteria, expert chemists performed systematic validation of routes. To calibrate the scores to align with the established qualitative criteria, we developed an augmented scoring system with a small-data fine-tuning stage on expert labels. Concretely, starting from the pretrained DeepSets encoder backbone~$\phi$ and scorer~$\sigma$ (trained to predict TED), we use a lightweight, expert-supervised adaptation using low-rank adapters (LoRA)~\cite{hu2022lora} on the encoder’s hidden layers and a new classification head (Figure \ref{fig:fine_tuned_model}). Specifically, the base encoder~$\phi$ weights and the scorer~$\sigma$ remain frozen. Given an expert label $y\in\{1,2,3,4,5\}$, we optimize only the LoRA parameters of $\phi$ and the new classification head by minimizing the cross-entropy. This design preserves the original distance predictor while allowing expert-specific adaptation with minimal data and parameters. The continuous TED regressor can still be reported as a secondary signal, but the decision-making is also driven by the rating to reflect expert preference directly.

\vspace{-1mm}
\section{Results}

We evaluate the feasibility rating prediction by comparing the predicted rating to expert-labeled rating. The code to reproduce our experiments is available at \url{https://github.com/yujiag21/Route-Scoring-with-Expert-Augmentation}.
Using 5-fold cross-validation on 120 expert-labeled routes, the model achieves the Spearman correlation coefficient of \textbf{0.78 ± 0.05} using the lowest reaction score as the route-level aggregate, and the Pearson correlation coefficient of \textbf{0.77 ± 0.06} using the average reaction score in the expert-annotated 5-point reaction rating scale. For downstream classification of routes into the three quality categories (Good / Plausible / Bad), we adopt the lowest reaction point as the route-level score and summarize the expert-augmented scoring scheme and corresponding per-class metrics in Table~\ref{tab:route_rating}. The overall accuracy of \textbf{67 ± 6.4\%}, with particularly strong performance in distinguishing between ``good'' and ``bad'' routes. 
The model's strengths lie in its reliable identification of good routes while maintaining high confidence in flagging truly problematic ones. 
\begin{table}[htpb]
\centering
\small
\caption{\textbf{Route rating and classification performance.} 
This table presents the score ranges used to define Good, Plausible, and Bad routes, along with the corresponding precision, recall, and F1-score of the classification.}
\label{tab:route_rating}
\begin{tabular}{lllll}
\toprule
Rating & Points & Precision & Recall & F1-score \\
\midrule
Bad        & 1,2 & $0.664 \pm 0.078$ & $\mathbf{0.860 \pm 0.116}$ & $0.738 \pm 0.015$ \\
Plausible  & 3,4 & $0.667 \pm 0.321$ & $0.393 \pm 0.088$ & $0.452 \pm 0.124$ \\
Good       & 5   & $0.781 \pm 0.134$ & $\mathbf{0.795 \pm 0.208}$ & $0.754 \pm 0.078$ \\
\bottomrule
\end{tabular}
\end{table}

\begin{figure}[htbp]
\centering
\includegraphics[width=\textwidth]{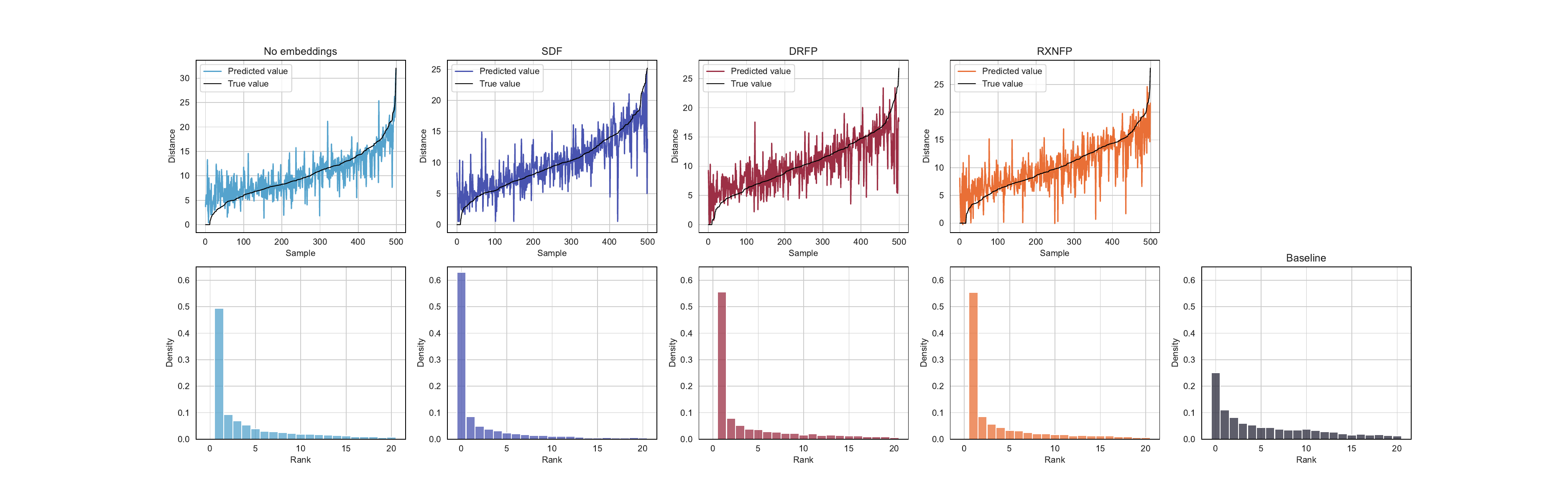}
\caption{\textbf{Model evaluation across different embedding methods.} Top: Line plots comparing predicted versus true distances for 500 randomly sampled test cases. 
Bottom: Top-k ranking performance histograms showing the probability of finding the true reference route within the model's k highest-ranked predictions (k$\leq$20). 
}
\label{fig: experiment_results}
\end{figure}

We evaluated the scoring on two key aspects: predicting route distances and ranking routes. For distance prediction, we used Mean Squared Error (MSE) and R-squared metrics (Table \ref{tab:performance_table}).
The SDF embedding outperformed all alternatives, achieving the lowest test MSE (7.102 ± 0.200) and highest R-squared (0.723 ± 0.003), explaining approximately 72\% of data variance.
To visualize prediction accuracy, we randomly sampled 500 test pairs of true and predicted distances for each method, sorted by true distance (Figure \ref{fig: experiment_results}, first row). The SDF embedding tracked true distances most consistently. 
SDF predictions showed the most balanced pattern across all distance ranges.
We also compared our model results with a tree-structured long short-term memory (tree-LSTM) \cite{mo2021evaluating} and showed significantly worse results 0.175.

\vspace{-1mm}

Besides distance prediction, we evaluated the model's ability to identify reference routes for novel molecules by comparing predicted top-1 routes against true reference routes. Table \ref{tab:performance_table} shows SDF achieved the best top-1 ranking accuracy of 0.603 ± 0.007.
The second row of Figure \ref{fig: experiment_results} visualizes the ratio of true top-1 routes appearing in predicted top-k routes across distance ranges. Figure \ref{fig:predited_ranking} in \ref{app:results} further details the proportion of reference routes within the top-20 predictions. SDF demonstrated superior performance here as well, identifying approximately 95\% of reference routes within its top-20 predictions.

\vspace{-1mm}
\begin{table}[h]
\small
    \centering
    \caption{\textbf{Performance of the scoring model.} 
    This table shows the mean and standard deviation ($\pm$) for mean squared error (MSE), R-squared, and top-1 ranking accuracy across different reaction embedding methods.}
    \label{tab:performance_table}

    \begin{tabular}{lccc}
        \toprule
        Embedding & MSE & R-Squared & Top-1 accuracy \\
        \midrule
        No embeddings & $8.683 \pm 0.362$ & $0.664 \pm 0.011$ & $0.442 \pm 0.004$ \\
        SDF           & $\mathbf{7.102} \pm \mathbf{0.200}$ & $\mathbf{0.723} \pm \mathbf{0.003}$ & $\mathbf{0.603} \pm \mathbf{0.007}$ \\
        DRFP          & $10.887 \pm 0.541$ & $0.566 \pm 0.006$ & $0.497 \pm 0.002$ \\
        RXNFP         & $11.318 \pm 0.243$ & $0.547 \pm 0.011$ & $0.488 \pm 0.007$ \\
        Baseline    & - & - & $0.175 \pm 0.002$ \\
        \bottomrule
    \end{tabular}
\end{table}

\vspace{-1mm}
\section{Conclusion}
We presented an expert-augmented framework that bridges the gap between data-driven machine learning and chemical domain knowledge.
The core of this system is a DeepSets-based model, pre-trained on large-scale patent data and fine-tuned on human assessments, which allows the system to capture implicit community standards. Consequently, the model delivers dual outputs: quantitative multi-objective route scores and interpretable qualitative feasibility ratings. This integrated approach not only offers a robust metric for route selection but also holds promise for other domains, such as healthcare, that require aligning AI with expert human judgment.

\bibliographystyle{unsrt}
\bibliography{science_template}


\clearpage

\appendix

\section{Data}

\subsection{Scoring metric}
\label{app:Scoring_metric}
To produce a universal and interpretable score for route quality, we adopted the \textbf{Tree Edit Distance (TED)} \cite{Bille2005ted, genheden2021ted} between any given route and the reference route as a simple yet robust metric. We operated under the assumption that the reference route represents the plausible synthetic pathway for each target molecule within its respective route set.

TED was selected after a comparative analysis of several candidate metrics, including the bond–atom–overlap–based similarity metric described in \cite{genheden2021ted}, which combines atom-level mapping and bond-formation similarity to capture both step order and key transformations. The details of this analysis are provided in \textit{Analysis of scoring metrics} in supplementary materials. Briefly, we first scored all routes using each metric, then identified and examined the 20 routes showing the largest score changes—both from high to low (“good-to-bad”) and from low to high (“bad-to-good”)—for each metric. We found that, although TED is conceptually simpler, it exhibited better generalizability, consistently distinguishing genuinely high-quality routes from poor ones. In contrast, more complex similarity metrics occasionally misclassified mediocre routes as poor (false negatives) or vice versa. By providing a quantitative measure of structural and strategic similarity between synthetic routes, TED served as a suitable prior for the evaluation score, and our scoring model was trained to predict the TED between the proposed and reference routes as the target. 

\subsection{Data collection}

To learn the criteria used to design the previously reported syntheses, we collected 47303 historical synthetic routes for various molecules from \textit{Journal of Medicinal Chemistry} as the reference routes for these molecules. Reference routes were constructed by first grouping reactions in the Reaxys database \cite{reaxys} by citation and then creating reaction networks from which routes could be extracted with a depth-first search \cite{mo2021evaluating}. Journal of Medicinal Chemistry, years 2000 - 2020, was used as the primary literature source in the Reaxys database. 

\begin{figure}[htbp]
\centering
\includegraphics[scale=0.5]{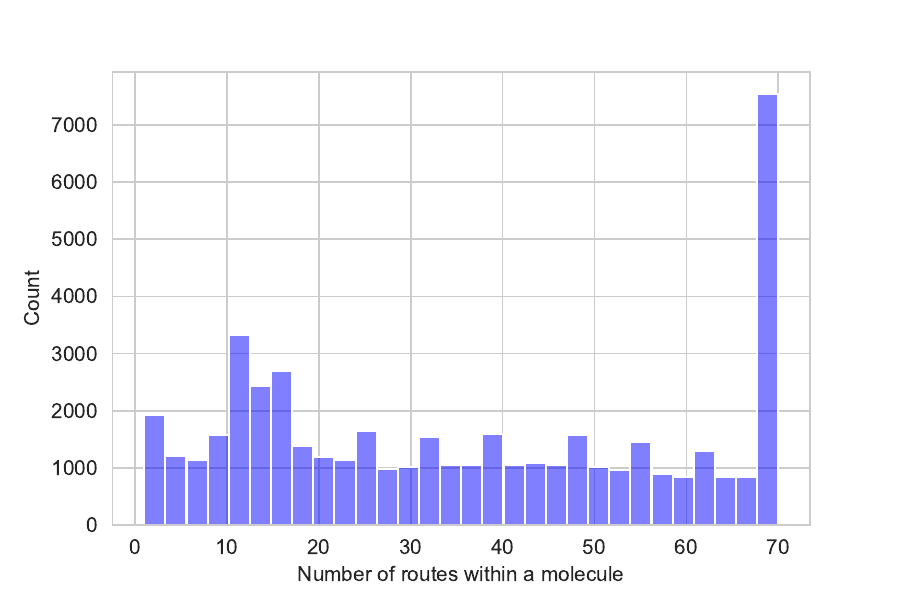}
\caption{\textbf{Distribution of the number of routes for each molecule.} We set the maximum number of output routes as 70 for each molecule during generation. }
\label{fig: route_count_distribution}
\end{figure}

We used AiZynthFinder to generate artificial alternative routes for each target molecule. The maximum number of output routes was set to 70, ensuring that up to the top 70 ranked routes were generated and selected by AiZynthFinder for each molecule. The full dataset consists of 1718004 routes, for 47303 different molecules. For the AiZynthFinder experiments, we used expansion and filter models trained on data from the US Patent and Trademark Office \cite{Genheden2020filter, genheden2023train}, and a stock of eMolecule downloaded in January 2023. 
On average, each target molecule has about 36 routes. 
Figure \ref{fig: route_count_distribution} shows the distribution of the number of routes for each molecule. From the figure, it can be seen that the most common number of routes per molecule is 70, while only 5 \% of all molecules have 5 or fewer routes associated with them.

\subsection{Data processing}

\label{section:feature}
We incorporated both route-level properties and reaction-level features, as well as the Morgan fingerprint \cite{rogers2010extended} of target molecules, as an input to the model.

\textbf{Route properties} include \textit{cost}, \textit{volume}, and \textit{structural complexity}. The \textit{cost} of a route is the sum of the molecule and reaction costs calculated using the RouteCostScorer function in AiZynthFinder, following the cost measurement by Badowski et al. \cite{badowski2019selection}.
The \textit{volume} is the number of intermediates that need to be synthesized in a route. The \textit{structural complexity} is measured as the total SCSscore \cite{coley2018scscore} value of these intermediates along the route. Volume measures the number of items involved in the synthesis process, whereas complexity evaluates the overall synthesis complexity of these items.

\textbf{Reaction features} consist of prior \textit{feasibility} score and molecular \textit{embedding} for each reaction and \textit{fingerprint} of target compound. To calculate reaction prior feasibility, a statistical analysis of the in-house reaction data subset was performed. Reactions were labeled ``successful'' if their yield was recorded as more than 20\% and labeled ``unsuccessful'' if their yield was recorded as zero along with a conclusion phrase ``Reaction failed'' \cite{{maloney2023negative}}. All labeled reactions were then grouped by the reaction class as identified by the NameRxn software \cite{{nameRxn}}. Each reaction class was given points for the total number of successful experiments and the average success rate of the reactions belonging to a certain reaction class. The specific point allocation system and evaluation criteria are shown in Table \ref{tab:points_allocation}.
As a result of this method, each individual reaction got a number between 0 and 6, depending on the reaction class to which it belongs.

In order to perform predictions on the reactions, we represented these reactions through embedding functions which take the SMILES string representation of a reaction as an input and output its embedding vectors. 
In our experiments, we tested three different methods of embedding: \textit{Structural Difference Fingerprint}(SDF)~\cite{schneider2015development},\textit{Chemical Reaction Fingerprint} (RXNFP)~\cite{schwaller2021mapping} and \textit{Differential Reaction Fingerprint} (DRFP)~\cite{probst2022reaction}, comparing them with a baseline model where we only use the reaction class feasibility as input for the route feasibility. 

\begin{table}[htbp]
    \centering
    \caption{\textbf{Prior scoring criteria for reaction classification.}
    This table defines how the number of prior experiments and observed success rates contribute to the assigned point score for a reaction.}
    \label{tab:points_allocation}
    \begin{tabular}{ccc}
        \hline
        Experiment Count & Success Rate & Points \\
        \hline
        $>5000$     & $>0.9$       & 3 \\
        $500$--$5000$ & $0.75$--$0.9$ & 2 \\
        $50$--$500$   & $0.6$--$0.75$ & 1 \\
        $<50$       & $<0.6$       & 0 \\
        \hline
    \end{tabular}
\end{table}

\section{Training Details}

We perform stratified $K$-fold cross-validation (default $K{=}5$) using quantile bins of the continuous TED distance for stratification. In each fold, we train with AdamW on the training split and select the seed/fold snapshot with the highest validation Spearman correlation between predicted classes and expert ranks. We report accuracy and macro-F1 on the validation split, together with Spearman/Pearson correlations between predicted classes and expert ranks, and (for monitoring only) the MSE between the frozen regression output and the ground-truth distance.


\section{Chemical Analysis}

The per-reaction approach was deemed reasonable because the molecular targets in this study are relatively small. As a result, strategic considerations such as synthetic convergence and cost of reaction order have a lesser impact. Consequently, the reaction order was only penalized when it raised functional group incompatibility issues. 

In order to further validate the best distance scoring model trained with SDF fingerprints, we have performed a systematic analysis of retrosynthesis predictions for 120 random molecules from the test set. 

We present a detailed description of the heuristic criteria, exemplified with selected representative cases: 
\begin{itemize}
    \item 5 points. In route shown in Figure \ref{fig: m_17} (molecule 17, predicted distance 1.46), the model succeeded in tracing back the 2,6-difunctionalized-pyridazinone target to three commercially available building blocks of the approximately same size, using two disconnections scored as highly feasible by experts.  In the forward synthetic sense, the model outcome suggested an N-arylation of 6-chloropyridazinone with 1-bromo-2,3-dichlorobenzene. This Ullman-type reaction can be expected to be feasible and chemoselective towards the aryl bromide moiety, and close precedent supports the assessment~\cite{liang2013rational}. The subsequent nucleophilic aromatic substitution at the 6-chloropyridazinone position with an alkyl alkoxide presents no evident issues as well~\cite{campayo2005new}.  It is worth mentioning that the corresponding originally observable route is conceptually distinct, i. e. features different disconnections.
    
    \begin{figure}[ht]
    \centering
    \includegraphics[width=0.5\textwidth]{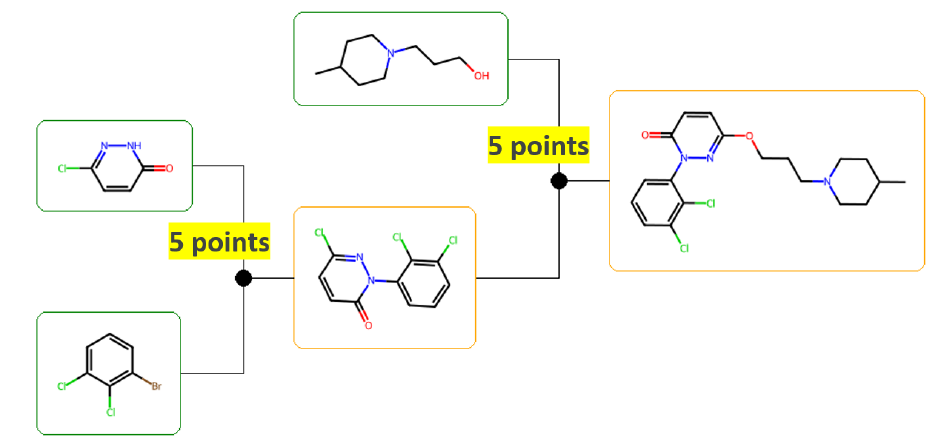}
    \caption{A predicted route for molecule 17 showing steps scoring 5 points.}
    \label{fig: m_17}
    \end{figure}

    \item 4 points. A step concerning the first case scenario is presented in the generated route for molecule 6 with a predicted distance of 5.5 (Figure \ref{fig:X2_a}). In the last step, the model suggested a carboxylic acid esterification using ethyl iodide---a quite reactive electrophile---in the presence of other nucleophilic sites (hydrazone and thiazole). Switching ethyl iodide for ethanol under acyl coupling conditions would be optimal and the overall route concept would remain unchanged.
    An example of a superfluous step scoring 4 points is found in route X3 in Figure \ref{fig:X2_b}. The final amidation should be feasible by using the proposed acyl fluoride directly \cite{campayo2005new}. 
    
    \begin{figure}[htbp]
        \centering
        \begin{subfigure}[b]{0.75\textwidth}
            \centering
            \includegraphics[width=\textwidth]{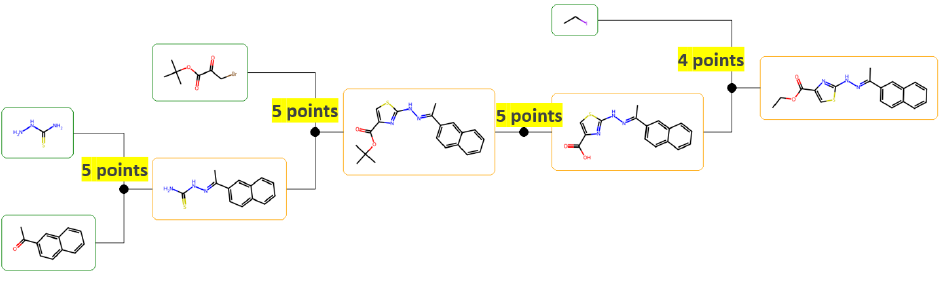}
            \caption{}
            \label{fig:X2_a}
        \end{subfigure}
        \hfill 
        \begin{subfigure}[b]{0.45\textwidth}
            \centering
            \includegraphics[width=\textwidth]{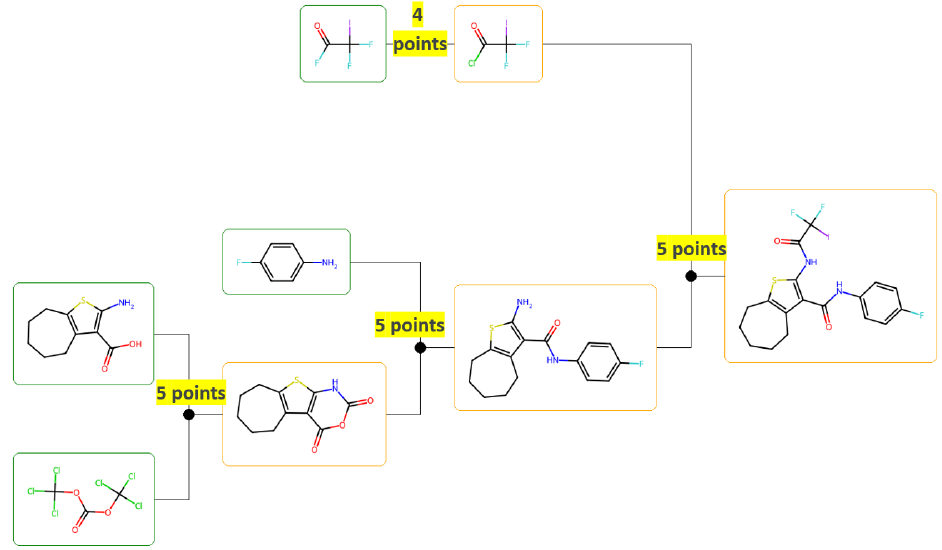}
            \caption{}
            \label{fig:X2_b}
        \end{subfigure}
        \caption{Routes containing 4-point steps.}
        \label{fig:X2}
    \end{figure}

    \item 3 points. The steps under this category would lead to a retrosynthetically correct route if an appropriate single protection group scheme enables that transformation and whose implementation does not alter the predicted route at a conceptual level. Such a protecting group strategy concerns any synthetic maneuver to eliminate and later reconstitute a reactive center, such as the use of blocking groups or circular redox manipulations. This scoring also applies to steps involving inadequate protecting groups. 
    For instance, route X4 would require protection of the primary alcohol at the propiolactamic starting material before the suggested N-acylation~\cite{mowery2007mimicry}, and subsequently a deprotection to claim the alcohol back. The conceptual synthetic strategy remains the same (Figure \ref{fig: X3}).
    \begin{figure}[htbp]
    \centering
    \includegraphics[width=0.5\textwidth]{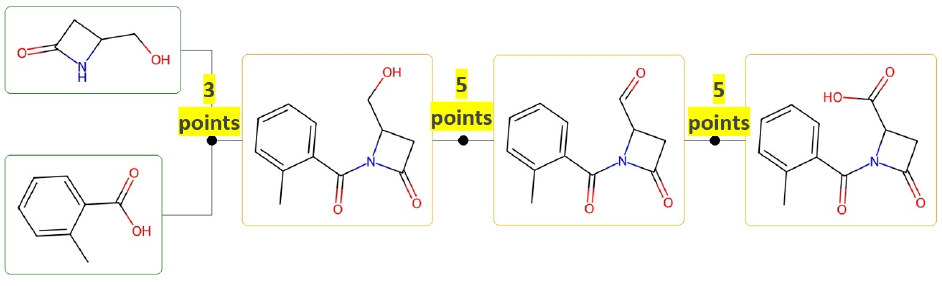}
    \caption{A predicted route containing a step scoring 3 points.}
    \label{fig: X3}
    \end{figure}
    
    \item 2 \& 1 points. To illustrate the last two heuristics, we selected the predicted route in Figure \ref{fig: X4}. In this sequence, one step was assessed at 2 points and another at 1 point. The route begins with a Chan–Lam N arylation, which can only be proposed with low confidence (2 points). Both N–H sites—the amidic and the anilinic—are similar under this reactivity manifold, and, to the best of our knowledge, there is no precedent for this transformation on such a polyheterocyclic substrate. A selective N protection would face the same limitation and therefore cannot be implemented with the information available at this stage. The second step requires a selective bromination of an electron-poor fluoroaniline fragment in preference to an electron-rich phenyl anilide fragment, which is highly unlikely. We therefore consider this transformation implausible, and it was scored 1 point.
    \begin{figure}[htbp]
    \centering
    \includegraphics[width=0.8\textwidth]{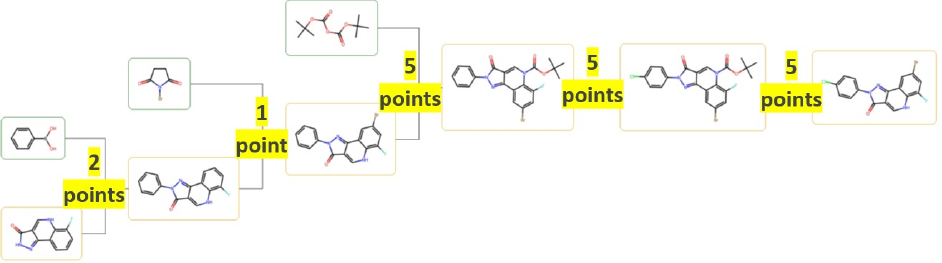}
    \caption{A predicted route containing steps scored 2 and 1 points..}
    \label{fig: X4}
    \end{figure}
\end{itemize}

\cite{US20030144303A1, gurovets1985hydrodehalogenation}.

\section{Results}
\label{app:results}

\begin{figure}[ht]
    \centering
    \includegraphics[scale=0.4]{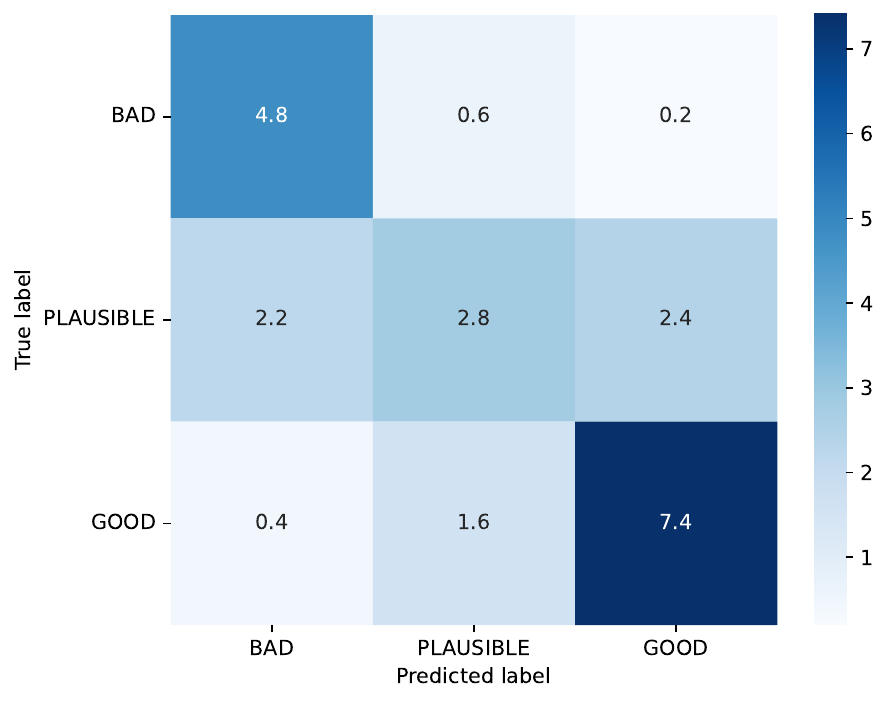}
    \caption{
        \textbf{Confusion matrix comparing model-generated route scores with expert chemist ratings.} The numbers in each cell represent the average count of routes falling into each category combination of 5 folds.  
}
    \label{fig: Confusion_Matrix}
    \end{figure}

Figure \ref{fig: Confusion_Matrix} presents the mean confusion matrix comparing route scores with expert ratings in a 5-fold cross-validation. 
The matrix demonstrates strong agreement between model and expert assessments, with most routes falling along the diagonal, particularly for good routes. However, there is some misclassification, notably in routes receiving middle-range scores where the model showed more uncertainty between the ``Good" and ``Plausible" categories.

Overall, the comparative analysis shows that the SDF embedding consistently outperforms other approaches. An interesting finding emerged when comparing the different embedding approaches: The more sophisticated embedding methods—DRFP and RXNFP—performed worse.
One reason that may explain the performance difference between the methods could lie in the architecture of the embeddings. SDF is simple and raw reaction representation, and thus less likely to be overfit or to learn unnecessary patterns. This analysis suggests that in the context of route distance prediction, simpler embedding approaches may be more effective than more sophisticated ones. The SDF method's straightforward approach to capturing structural changes in reactions appears to provide the right balance of expressiveness and simplicity for this specific task.

\begin{figure}[htbp]
\centering
\includegraphics[scale=0.4]{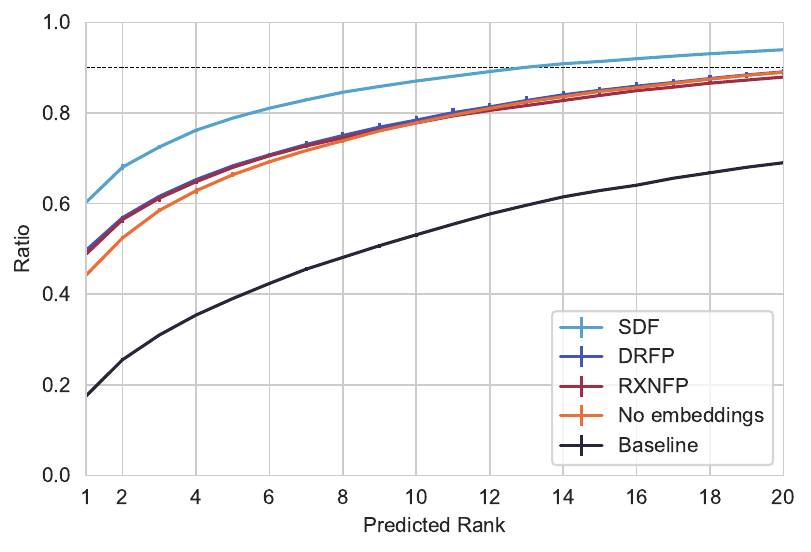}
\caption{\textbf{Proportion of reference routes in prediction.} The proportion of reference routes that is in the top 20 predicted routes with threshold line at 0.95. The error bars show the standard deviation.}
\label{fig:predited_ranking}
\end{figure}
\vspace{-1mm}

\section{Related Works}

Research on chemical route evaluation is still relatively limited. Butters et al \cite{butters2006critical} proposed general evaluation criteria named SELECT, including Safety, Environmental, Legal, Economic, Control, and Throughput. These criteria are rather theoretical and rough. The main problem is that this evaluation lacks precise quantification for some of the criteria and a way to combine these criteria into a single score, making it hard to balance them when designing or evaluating routes. 

Several algorithmic methods have been developed for generalizable and quantifiable synthesis evaluation. Schwaller et al. \cite{schwaller2020predicting} introduced four key metrics—coverage, class diversity, round-trip accuracy, and Jensen–Shannon divergence—to assess single-step retrosynthesis models. Lin et al. \cite{lin2020automatic} used a heuristic scoring function based on the F1 score of common scaffolds between target and starting molecules, while Badowski et al. \cite{badowski2020synergy} developed a neural network-based scoring model that predicts reaction probabilities from Morgan fingerprints. These methods, however, focus only on single-step evaluations without considering route-level information. Mo et al. \cite{mo2021evaluating} argued that heuristic scoring may overlook creative retrosynthesis routes and developed a model to assess the strategic viability of multi-step synthesis pathways using an LSTM encoder. Their model ranks routes based on strategic value, focusing on pathway-level relationships. In contrast, our work incorporates both single-step reaction assessments and overall route properties. A further difference is that their model was trained as a ranking task where absolute score values are less meaningful, whereas our scoring model aims to provide a universal and interpretable assessment standard based on score levels.
    
\section{Discussion}

In this work, the scoring model treats reactions within a synthetic route as an unordered set. 
By treating reactions as a set, we established a baseline for what can be achieved without considering reaction ordering. Our choice of permutation invariance prioritizes simplicity and enables the model to focus on reaction-level properties independently of sequence order. However, we acknowledge that the order of reactions within a route carries potentially informative context. This study explores how effectively routes can be evaluated when treated as unordered sets, and we raise the possibility in future work that incorporating reaction order could further improve the predictive performance of the model.

Moreover, while our model shows promising performance overall, it encounters some difficulty distinguishing between adjacent categories. Future work could explore more nuanced scoring approaches through human-in-the-loop learning that account for multiple reference routes or incorporate other metrics of synthetic efficiency beyond structural similarity. By incorporating granular expert chemists’ feedback in an interactive way for route evaluations, the system could learn to recognize nuances in synthetic strategy that go beyond simple structural similarity to reference routes. potentially leading to a more comprehensive evaluation framework that better reflects the multifaceted nature of route optimization in real-world synthesis planning.  This interactive learning approach could help refine both the TED-based scoring and classification boundaries while capturing the tacit knowledge that experienced chemists use to evaluate routes - considerations that may not be apparent from literature references or score scale augmentation alone.

\end{document}